# Large-Scale YouTube-8M Video Understanding with Deep Neural Networks

Manuk Akopyan
Institute for System Programming
ispras.ru
manuk@ispras.ru

Eshsou Khashba
Institute for System Programming
ispras.ru
eshsou@ispras.ru

**Abstract**

*Video classification problem has been studied many years. The success of Convolutional Neural Networks (CNN) in image recognition tasks gives a powerful incentive for researchers to create more advanced video classification approaches. As video has a temporal content Long Short Term Memory (LSTM) networks become handy tool allowing to model long-term temporal clues. Both approaches need a large dataset of input data. In this paper three models provided to address video classification using recently announced YouTube-8M large-scale dataset. The first model is based on frame pooling approach. Two other models based on LSTM networks. Mixture of Experts intermediate layer is used in third model allowing to increase model capacity without dramatically increasing computations. The set of experiments for handling imbalanced training data has been conducted.*

1. **Introduction**

In last few years image classification problem is enjoying a renaissance with an arise of deep learning approach. Many models have been designed (AlexNet [1], VGGNet [2], ResNet [3], Inception [4]), which effectively recognize image content. The first reason of that is publication of the free large-scale data base of high resolution images ImageNet [5], and usage of highly effective operations on GPU-s providing high-throughput computing on relatively cheap cost. Prediction accuracy of such deep learning models approaches to human level performance.

The next step in this direction is not just object recognition in static images but an action recognition, video classification. There are few benchmarks providing input datasets for these problems (Sports-1M [6], UCF-101[7] etc). Recently YouTube-8M benchmark [8] was published with dataset size exceeding competitors' dramatically. As like as in object recognition in this area many hand-crafted approaches to video-frame feature extraction, such as Histogram of Oriented Gradients (HOG), Histogram of Optical Flow (HOF), Motion Boundary Histogram (MBH) around spatio-temporal interest points [9], in a dense grid [10], SIFT [11], the Mel-Frequency Cepstral Coefficients (MFCC) [12], the STIP [13] and the dense trajectories [14] existed. Set of video-frame features then encoded to video-level feature with bag of words (BoW) approach. The problem with BoW is that it uses only static video-frame information disposing of the time component, the frame ordering. Recurrent Neural Networks (RNN) show good results in modeling with time-based input data. A few papers [15, 16] describe solving video classification problem using Long Short-Term Memory (LSTM) networks and achieve good results.

This paper describes three models used to solve video classification problem for YouTube-8M. Described models were used in Google Cloud & YouTube-8M Video Understanding Challenge [17]. First model is based on BoW: time-based frame codes pooled then classified, second and third model based on LSTM approach. In contrast with [15, 16], we used also video soundtrack information as provided by YouTube-8M.

The paper is organized as follows. Section 2 reviews related works of video classification problem using deep learning. Section 3 presents brief overview of YouTube-8M dataset. Section 4 describes the proposed deep learning models to solve multi-label multi-class classification on YouTube-8M dataset. Section 5 provides results of training proposed models to dataset. Finally, section 6 concludes the paper by summarizing the main points addressed through this paper.

2. **Related works**

Video classification problem has been studied many years. Many approaches solving the problem has been developed using hand-crafted features.

GPU progress and winning ImageNet competition by Krizhevsky et al. [1] made deep learning approach more popular. At this moment Inception-v3 network gets ~3%

top-5 error. Deep learning approach used for video classification was described in recent papers [6, 15, 16]. Karpathy et al. [6] proposed use a ConvNet approach based models on Sports-1M, UCF-101 benchmarks. Ng et al. [15] proposed few feature pooling architectures and LSTM architectures on Sports-1M, UCF-101 benchmarks. Abu-El-Haija et al. [16] worked on YouTube-8M, Sports-1M benchmarks. They provided few baseline approaches (Deep Bag of Frames, LSTM). They used frame-level features and aggregated video-level features in their models.

YouTube-8M contains encoded features in contrast with Sports-1M, UCF-101 datasets where various pre-processing approaches can be applied on a raw video frame images. Researchers in [15, 16] do not use audio information. In public version of YouTube-8M encoded video and audio feature for each frame are provided, preventing any pre-processing technique on raw data.

Shazeer et al. [18] present intermediate the MoE layer, which increase model capacity without a proportional increase in computation. In the proposed approach the MoE layer is stacked between LSTM layers. The presented model was applied to the tasks of language modeling and machine translation and competitive results have been achieved.

In this work the MoE layer in combination of LSTM layers was used for video classification problem.

## 3. Input Data

The YouTube-8M benchmark contains 4716 classes and more than 8 million videos. The benchmark is split into three subsets: Train (~5,7M), Validate (~1,6M) and Test (~0.8M). YouTube-8M is available in two datasets: the frame-level features dataset and the video-level features dataset.

**Frame-level features dataset**. Original videos have been preprocessed to extract frame-level features. Each video is decoded at 1 fps up to the first 360 seconds and then decoded frames are feed into the Inception-v3 network. Features vector of length 2048 is taken before classification layer. To reduce feature dimension to 1024 PCA and quantization is applied.

Also, audio features are extracted from videos [20] and added to the dataset.

The total size of frame-level dataset is about 1,7TB. Extracted features are stored in tfrecords format and are available on Internet [8]. Each video record has the following structure:
*context[*
  *video_id*
  *label_list*
*]*
*feature[*
  *rgb_list*
  *audio_list*
*]*

**Video-level features dataset**. Features in this dataset are aggregated from the frame-level features dataset.

In our models we are going to use the frame-level dataset only.

## 4. Models

This section provides description of the models used to train and predict themes of video. The first model based on Bag-of-Frames approach – mini-batch of input video-frame features pooled along time axis to get video-level features. This allows to model static spatial information over time axis.

As input data has time axis (time based) we decided to use RNN, which allows extraction of the temporal information of a sequential input data. The second model presents a network with few LSTM layers and classifier. The third model also is RNN based, but here we add intermediate MoE layer based on [18, 19].

### 4.1. Bag-of-Frames architecture

Bag-of-words (or as in our case Bag-of-Frames) representation is widely used in the video classification problem [15, 16, 22, 23]. Each input sample corresponds to a video, has a set of video-labels, and a sequence of frame features. Frame feature could be a hand-crafted feature for each input video-frame or as in our case a raw video-frame encoded by Inception-v3.

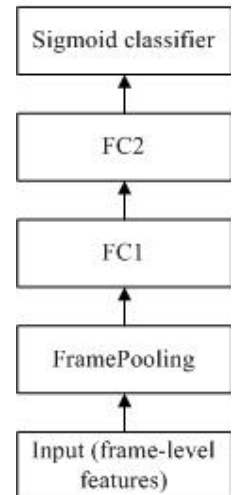

Figure 1. Bag-of-Frames architecture.

Bag-of-Frames architecture is illustrated in Figure 1. For each sample in a training dataset there is a set of frame-level features and ground-truth video-level labels. We need to train model to predict video-level labels. Input data (*batch_size, max_frames, feature_size*) is sent to FramePooling layer, where pooling between time frames of each sample is applied. As in [15] we use max-pooling to get one feature-vector from all time-based frame-level features of each input sample. After FramePooling layer two FC layers are used. And on the top level we use sigmoid classifier.

Batch Normalization layer is added after Input, FC1, and FC2 layers to provide stability, convergence speed and some regularization. Also dropout is applied to the output of Input and FC1 layers (with probability 0.3). This model is trained with 90 frames for each input sample (*max_frames*=90) in time dimension. Sigmoid cross entropy loss is used to train multi-class multi-label classifier. RMSProp with decay rate 0.9 is used as optimizer. We use base learning rate $10^{-4}$, which decay after each $2*10^7$ samples.

### 4.2. Simple_LSTM architecture

In Bag-of-Frames architecture frame-level features are pooled (max) along time axis, and hence we are loosing frame ordering information. In this section we consider simple LSTM architecture, which allows to model long-term temporal information. LSTM networks may connect previous information to the present task (e.g. connect information from the previous frames of a video to the current frame), they are capable of learning long-term dependencies.

In Figure 2 Simple_LSTM architecture is presented. We use *N* LSTM layers, each layer has *hidden_size* units. Configurations with *N*=[2, 3], *hidden_size*=[1024, 2048] have been tested. *N* LSTM Layers are stacked so that output of (*i-1*)-th layer is input for the *i*-th layer. Stacked LSTM network is unrolled *max_frames* times. Weighted sum of outputs of LSTM_N layer is sent to sigmoid classifier:

$$Sigmoid\_input = \sum_{i=1}^{max\_frames} w_i * out_i \quad (1)$$

where $out_i$ is unrolled LSTM_N layers *i*-th output, $w_i$ is a weight:

$$w_i = \frac{i}{max\_frames} \quad (2)$$

Therefore the output of the first time frame gets minimal weight and the output of the last time frame gets maximal weight.

Batch Normalization layer is added after Input layer to provide stability, convergence speed, and some regularization. Dropout is applied to the output of each LSTM layer (with probability 0.4). In some experiments residual connections have been used. We trained this model with *max_frames*=90 in time dimension. Sigmoid cross entropy loss is used to train multi-class multi-label classifier. RMSProp with decay rate 0.9 is used as optimizer. We used base learning rate 2*10-4, which decay after each $10^7$ samples.

### 4.3. LSTM_MoE architecture

The idea of the Mixture of Experts [19] is to train a set of experts, where each expert specializes on a subset of cases. To choose expert for each sample gating network is designed. The output of gating network is softmax layer, which provides probability of choosing particular expert (number of output probabilities is equal to number of experts).

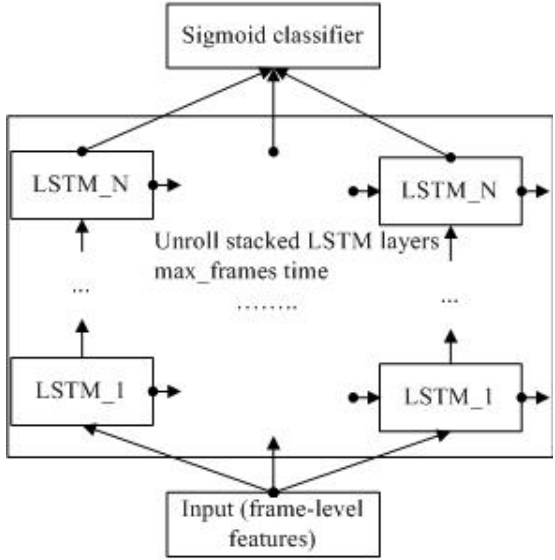

Figure 2. Simple_LSTM architecture.

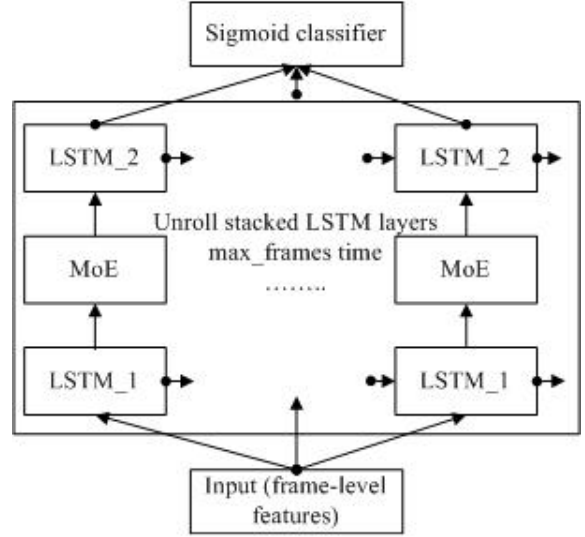

Figure 3. Lstm_MoE architecture.

In early works [19, 24, 25, 26] the MoE was used as a top layer. Shazeer et al. [18] proposed to use MoEs as a general purpose neural network component. We implemented MoE layer in Tensorflow library [27] and used it in the third model.

In Figure 3 LSTM_MoE architecture is presented. LSTM_MoE is similar to Simple_LSTM except for the following details:
- only 2 LSTM layers are used
- *hidden_size*=512
- output of the LSTM_1 layer is input for the MoE layer, output of the MoE layer is input for the LSTM_2 layer.

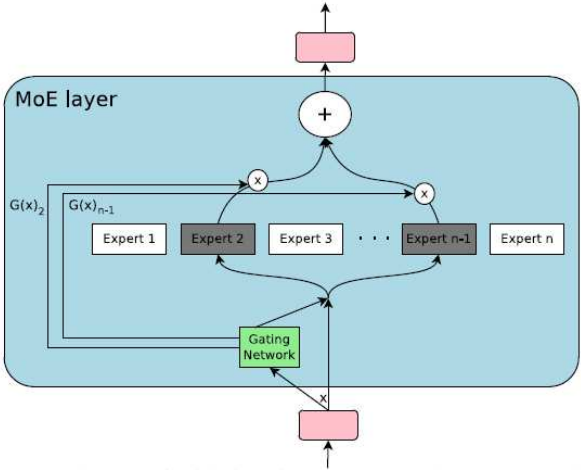

Figure 4. MoE layer.

In Figure 4 schematic presentation of MoE layer is provided. Output from the LSTM_1 layer is sent to gating network. Gating network selects a sparse combination of the experts to process each input sample. This kind of sparse selection allows saving computations.

MoE layer consists of Gating network and a set of *n* Expert networks. We used *n*=64 Expert networks, 4 of which are active for each sample. Each Expert network consists of two FC layers with 1024 hidden units.

Gating network tends to choose the same experts. To provide load balancing two loss functions $Loss_{importance}$, $Loss_{load}$ are provided which are used to penalize such a behavior.

$$Importance(X) = \sum_{x \in X} G(x) \qquad (3)$$

$$Loss_{importance}(X) = w_{importance} * CV(Importance(X))^2 \quad (4)$$

where $G(x)$ is output of gating network, CV is coefficient of variation of the set of importance values, and $w_{importance}$ is hyper-parameter.

$$Load(X) = \sum_{x \in X} P(x) \qquad (5)$$

$$Loss_{load}(X) = w_{load} * CV(Load(X))^2 \qquad (6)$$

where $P(x, i)$ is the probability that $G(x)_i$ is nonzero, given a new random choice of noise on element *i*, but keeping the already-sampled choices of noise on the other elements, CV is coefficient of variation of the set of load values, and $w_{load}$ is hyper-parameter.

LSTM_MoE is a very memory consuming model, so only limited experiments with this model have been done.

## 5. Experiments

In this section we report the results of experiments on YouTube-8M dataset. We compare the results of training BoF, Simple_LSTM, LSTM_MoE on the large-scale YouTube-8M frame-level dataset. All models have been implemented on the Tensorflow library.

For the evaluation we use *Hit@1* and *Global Average Precision (GAP)* metrics.

**Hit@1**. This is the fraction of test samples where the top-1 prediction contains in the ground truth labels of the sample.

*Hit@1(X)*=

$$\frac{1}{|X|} \sum_{x \in X} labels(x)[\,argmax(predictions(x))] \quad (7)$$

where *X* is a testing set of samples, *labels(x)* is a vector of ground truth labels (0 or 1), *predictions(x)* are predicted labels, *argmax(vec)* is an index of maximum value in the vector *vec*.

**GAP.** Definition of *GAP* corresponds to the one provided on competition site [17]. Let *N* be a number of predictions, *p(i)* is the precision, and *r(i)* is the recall.

$$GAP = \sum_{i=1}^{N} p(i) * \Delta r(i) \qquad (8)$$

| Model | Hit@1 | GAP | Step number |
|---|---|---|---|
| BoF_4096 | 0.78 | 0.66 | 60k |
| Simple_LSTM_residual | 0.82 | 0.75 | 200k |
| Simple_LSTM_2048 | 0.85 | 0.77 | 100k |
| Simple_LSTM_l3_2048 | 0.87 | 0.79 | 80k |
| LSTM_MoE_512 | 0.8 | 0.76 | 50k |
| Frame-level_LSTM | 0.645 | --- | --- |
| Video-level_MoE | 0.633 | --- | --- |

Table 1. Results of various configurations on YouTube-8M.

**Results**. Table 1 contains the best results of experiments on BoF, Simple_LSTM, LSTM_MoE models. For all experiments in Table 1 the inputs are frame-level features (rgb, audio) with sizes 1024 and 128,

max_frames is 90.

Very often the first few seconds of video are not very informative in terms of video classification as they contain some text or intro animation which does not help in classification. So we skip first few seconds (frames) of the input data. This allows to improve evaluation metrics by ~0.6%. In all our experiments the first 20 frames of input data are skipped.

BoF_4096 represents the BoF model. The number of hidden units in FC1 and FC2 is 4096. The batch size is 512. The model was trained for 60k steps (~11 epochs) on Tesla K40 GPU. The training time was 2 days.

Simple_LSTM_residual represents the Simple_LSTM model. The number of hidden units in LSTM_1 and LSTM_2 is 1024. For both LSTM layers residual connection is used. The batch size is 256. The model was trained for 200k steps (~10 epochs) on Quadro P6000 GPU. The training time was 7 days.

Simple_LSTM_2048 represents the Simple_LSTM model. The number of hidden units in LSTM_1 and LSTM_2 is 2048. The batch size is 256. The model was trained for 100k steps (~5 epochs) on Tesla K40 GPU. The training time was 8 days.

Simple_LSTM_l3_2048 represents the Simple_LSTM model. Three LSTM layers are used in this configuration. The number of hidden units in each LSTM layer is 2048. The batch size is 256. The model was trained for 80k steps (~4 epochs) on Tesla K40 GPU. The training time was 13 days.

LSTM_MoE_512 represents the LSTM_MoE model. The number of hidden units in LSTM_1 and LSTM_2 is 512. We used n=64 Expert networks, 4 of which are active for each sample. Each Expert network consists of two FC layers with 1024 hidden units. The batch size is 128. The model was trained for 50k steps (~2 epochs) on Quadro P6000 GPU. The training time was 3 days.

Also some configurations from Table 1 were trained for additional time steps (+100k) but without essential improvements. The network just plateaus for each of the provided model configurations.

Frame-level_LSTM and Video-level_MoE shows the best performance in [16]. Our models improve baseline performance (Hit@1 metric) for 14-23%.

| Label_id | Sample_number | Sample_number (%) |
|---|---|---|
| 1 | ~860k | ~0.052 |
| 2 | ~680k | ~0.041 |
| 3 | ~520k | ~0.031 |
| … | … | … |
| 2100 | 501 | 3.01e-05 |
| … | … | … |
| 4715 | 99 | 5.95e-06 |
| 4716 | 100 | 6.01e-06 |

Table 2. Label counts in training set.

The training set ground-truth label analysis showed that labels are heavily unbalanced. Table 2 provides number of labels in training set. The ratio of the most frequent and least frequent label counts is 8.6k.

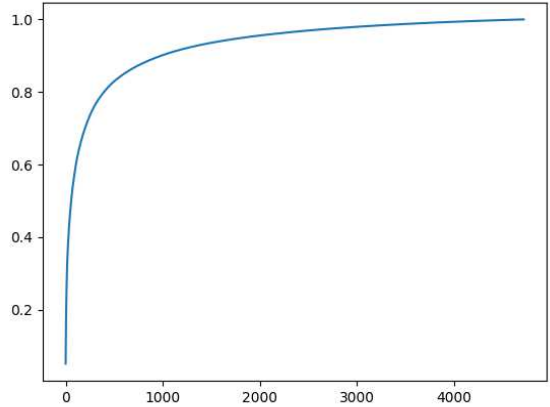

Figure 5. Sum of label numbers (in %) in training set.

In Figure 5 x-axis corresponds to label id and y-axis corresponds to sum of label counts in percents. 95% of all labels in training set compose of first 1834 labels.

One of the possible reasons of models plateau may be the label imbalance: networks just learn well for few top classes and for the most input samples they predict label from that top set of labels (majority classes).

In the next attempt to handle imbalanced training data we use penalized loss function. We put additional cost on the model for making classification mistakes on the minority class during training. We penalize false negative cases in loss function giving penalty coefficient inversely to number of labels in training set.

| Model | Hit@1 | GAP | Step number |
|---|---|---|---|
| Simple_LSTM_3_2048_bal | 0.73 | 0.61 | 80k |
| Simple_LSTM_2048_bal | 0.71 | 0.59 | 80k |
| LSTM_MoE_512_bal | 0.57 | 0.32 | 50k |

Table 3. Results of penalized model performance.

Unfortunately the penalized models converge slowly and need more time to train. In Table 3 the results of experiments working with the penalized models are provided.

Simple_LSTM_3_2048_bal represents penalized model Simple_LSTM. The number of LSTM layers is 3, the number of hidden units in LSTM layers is 2048. The model was trained with max_frames=120 in time dimension. The batch size is 128. The model was trained for 80k steps (~2 epochs) on Tesla K80 GPU. The training time was 2 days.

Simple_LSTM_2048_bal represents penalized model Simple_LSTM. Hyper-parameters are the same as for Simple_LSTM_2048. The model was trained for 80k steps (~4 epochs) on Tesla K40 GPU. The training time was 6 days.

LSTM_MoE_512_bal represents penalized model LSTM_MoE. Hyper-parameters are the same as for LSTM_MoE_512. The model was trained for 50k steps (~2 epochs) on Quadro P6000 GPU. The training time was 3 days.

Because of lack of time we have trained just three configurations and have not well tuned these set of experiments (tuning of these setups planed as future works).

The results in Table 3 are lower than in Table 1, perhaps more time and/or tuning of hyper-parameters will give better results. Also other approaches to handle with imbalanced classes (under-sampling majority classes, over-sampling minority classes etc) are to be applied.

6. **Conclusion**

We have presented three deep learning models for video classification. All provided models were trained on frame-level input data from the YouTube-8M dataset. The first model (BoF) used frame pooling approach. Simple_LSTM and LSTM_MoE used LSTM layers for the long-term temporal dependency modeling. In LSTM_MoE model Mixture of Experts layer have been implemented and trained.

In the input dataset we skip the first few seconds (frames) as they are usually useless in video classification tasks.

Our best model configurations improve baseline performance (Hit@1 metric) for 14-23%.

Also we have conducted the set of experiments for handling imbalanced training data by using penalized loss function. This leads to more distributed prediction results but drops overall *Hit@1* and *GAP* metrics. These experiments are a subject for further tuning and improvements.

**References**

1. A. Krizhevsky, I. Sutskever, and G. E. Hinton. Imagenet classification with deep convolutional neural networks. In Advances in neural information processing systems, pages 1097–1105, 2012.
2. K. Simonyan and A. Zisserman. Very deep convolutional networks for large-scale image recognition. arXiv preprint arXiv:1409.1556, 2014.
3. K. He, X. Zhang, S. Ren, and J. Sun. Deep residual learning for image recognition. arXiv preprint arXiv:1512.03385, 2015.
4. C. Szegedy, V. Vanhoucke, S. Ioffe, J. Shlens, and Z. Wojna. Rethinking the inception architecture for computer vision. In Proceedings of the IEEE Conference on Computer Vision and Pattern Recognition, pages 2818–2826, 2016.
5. J. Deng, W. Dong, R. Socher, L.-J. Li, K. Li, and L. Fei-Fei. ImageNet: A Large-Scale Hierarchical Image Database. In CVPR09, 2009.
6. A. Karpathy, G. Toderici, S. Shetty, T. Leung, R. Sukthankar, and L. Fei-Fei. Large-scale video classification with convolutional neural networks. In IEEE Conference on Computer Vision and Pattern Recognition (CVPR), pages 1725–1732, Columbus, Ohio, USA, 2014.
7. K. Soomro, A. R. Zamir, and M. Shah. UCF101: A dataset of 101 human actions classes from videos in the wild. In CRCV-TR-12-01, 2012.
8. YouTube-8M Dataset. https://research.google.com/youtube8m/.
9. I. Laptev, M. Marszaek, C. Schmid, and B. Rozenfeld. Learning realistic human actions from movies. In Proc. CVPR, pages 1–8, Anchorage, Alaska, USA, 2008.
10. H. Wang, M. M. Ullah, A. Klser, I. Laptev, and C. Schmid. Evaluation of local spatio-temporal features for action recognition. In Proc. BMVC, pages 1–11, 2009.
11. D. G. Lowe. Distinctive image features from scale-invariant keypoints. IJCV, 2004.
12. Z. Xu, Y. Yang, I. Tsang, N. Sebe, and A. Hauptmann. Feature weighting via optimal thresholding for video analysis. In ICCV, 2013.
13. I. Laptev. On space-time interest points. IJCV, 2007.
14. H. Wang and C. Schmid. Action recognition with improved trajectories. In ICCV, 2013.
15. J. Y.-H. Ng, M. J. Hausknecht, S. Vijayanarasimhan, O. Vinyals, R. Monga, and G. Toderici. Beyond short snippets: Deep networks for video classification. In IEEE Conference on Computer Vision and Pattern Recognition (CVPR), pages 4694–4702, 2015.
16. Sami Abu-El-Haija, Nisarg Kothari, Joonseok Lee, Paul Natsev, George Toderici, Balakrishnan Varadarajan, Sudheendra Vijayanarasimhan. YouTube-8M: A Large-Scale Video Classification Benchmark. arXiv:1609.08675 (2016).
17. Google Cloud & YouTube-8M Video Understanding Challenge. https://www.kaggle.com/c/youtube8m/.
18. Noam Shazeer, Azalia Mirhoseini, Krzysztof Maziarz, Andy Davis, Quoc Le, Geoffrey Hinton, Jeff Dean. Outrageously Large Neural Networks: The Sparsely-Gated Mixture-of-Experts Layer. arXiv preprint arXiv:1701.06538.
19. Jacobs, R. A., Jordan, M. I., Nowlan, S. J., and Hinton, G. E. 1991. Adaptive mixtures of local experts. Neural Comp. 3, 79-87.
20. Shawn Hershey, Sourish Chaudhuri, Daniel P. W. Ellis, Jort F. Gemmeke, Aren Jansen, Channing Moore, Manoj Plakal, Devin Platt, Rif A. Saurous, Bryan Seybold, Malcolm Slaney, Ron Weiss, Kevin Wilson. CNN Architectures for Large-Scale Audio Classification. International Conference on Acoustics, Speech and Signal Processing (ICASSP), IEEE (2017).
21. S. Ji, W. Xu, M. Yang, and K. Yu. 3d convolutional neural networks for human action recognition. In ICML, 2010.
22. K. Simonyan and A. Zisserman. Two-stream convolutional networks for action recognition in videos. In NIPS, 2014.

23. Bangpeng Yao, Dirk Walther, Diane Beck, and Li Fei-fei. Hierarchical mixture of classification experts uncovers interactions between brain regions. In NIPS. 2009.
24. Rahaf Aljundi, Punarjay Chakravarty, and Tinne Tuytelaars. Expert gate: Lifelong learning with a network of experts. CoRR, abs/1611.06194, 2016. URL http://arxiv.org/abs/1611.06194.
25. Ekaterina Garmash and Christof Monz. Ensemble learning for multi-source neural machine translation. In staff.science.uva.nl/c.monz, 2016.
26. Martín Abadi, Ashish Agarwal, Paul Barham, Eugene Brevdo, Zhifeng Chen, Craig Citro, Greg S. Corrado, Andy Davis, Jeffrey Dean, Matthieu Devin, Sanjay Ghemawat, Ian Goodfellow, Andrew Harp, Geoffrey Irving, Michael Isard, Yangqing Jia, Rafal Jozefowicz, Lukasz Kaiser, Manjunath Kudlur, Josh Levenberg, Dan Mane, Rajat Monga, Sherry Moore, Derek Murray, Chris Olah, Mike Schuster, Jonathon Shlens, Benoit Steiner, Ilya Sutskever, Kunal Talwar, Paul Tucker, Vincent Vanhoucke, Vijay Vasudevan, Fernanda Viegas, Oriol Vinyals, Pete Warden, Martin Wattenberg, Martin Wicke, Yuan Yu, Xiaoqiang Zheng. TensorFlow: Large-Scale Machine Learning on Heterogeneous Distributed Systems. arXiv:1603.04467 (2016).